\PassOptionsToPackage{dvipsnames}{xcolor}
\documentclass[anon,12pt]{colt2026} %

\title[Lookahead identification in adversarial MAB]{Lookahead identification in adversarial bandits:\\
accuracy and memory bounds}

\usepackage{times}

\coltauthor{%
 \Name{Nataly Brukhim} \Email{nbrukhim@princeton.edu}\\
 \addr DIMACS, Rutgers University\\
  \addr Institute for Advanced Study
 \AND
 \Name{Nicolò Cesa-Bianchi} \Email{nicolo.cesa-bianchi@unimi.it}\\
 \addr Università degli Studi di Milano\\
 \addr Politecnico di Milano%
 \AND
 \Name{Carlo Ciliberto} \Email{c.ciliberto@ucl.ac.uk}\\
 \addr Centre for artificial intelligence\\
 \addr University College London%
}

\usepackage{latexsym,color,graphicx, mathtools}
\usepackage{algorithm, algorithmic}
\usepackage[normalem]{ulem}
\usepackage[bf]{caption}

\newenvironment{proofof}[1]{\begin{proof}\textbf{of {#1}}}{\end{proof}}

 \usepackage{cleveref}   %

 \DeclarePairedDelimiter\ceil{\lceil}{\rceil}
\DeclarePairedDelimiter\floor{\lfloor}{\rfloor}

\def\A{{\mathcal A}}

\def\R{{\mathcal R}}

\def\E{{\mathbb E}}

\renewcommand{\paragraph}[1]{~\newline\noindent{\bfseries #1.}}

\newtheorem{claim}[theorem]{Claim}

\begin{document}

\maketitle

 \begin{abstract}
We study an identification problem in  multi-armed bandits. In each round a learner selects one of $K$ arms and observes its reward, with the goal of eventually identifying an arm that will perform best at a {\it future} time. In adversarial environments, however, past performance may offer little information about the future, raising the question of whether meaningful identification is possible at all.

In this work, we introduce \emph{lookahead identification}, a task in which the goal of the learner is to select a future prediction window and commit in advance to an arm whose average reward over that window is within $\varepsilon$ of optimal.  Our analysis characterizes both the achievable accuracy of lookahead identification and the memory resources required to obtain it. From  an accuracy standpoint, for any horizon $T$ we give an algorithm achieving $\varepsilon = O\bigl(1/\sqrt{\log T}\bigr)$ over $\Omega(\sqrt{T})$ prediction windows. This demonstrates that, perhaps surprisingly, identification is possible in adversarial settings, despite significant lack of information. We also prove a near-matching lower bound  showing that $\varepsilon = \Omega\bigl(1/\log T\bigr)$ is unavoidable. We then turn to investigate the role of memory in our problem, first proving that any algorithm achieving nontrivial accuracy requires $\Omega(K)$ bits of memory. Under a natural \emph{local sparsity} condition, we show that the same accuracy guarantees can be achieved using only poly-logarithmic memory.

Finally, we contrast lookahead identification with the more standard objective of regret minimization. We show a sharp separation between identification and regret: despite strong memory lower bounds for identification, sublinear regret can be attained using only limited memory. Specifically, we give an algorithm achieving $\tilde{O}(T^{2/3}K^{1/3})$ regret using poly-logarithmic memory. Our result significantly improves upon prior work both in terms of the regret and the memory bounds. 
\end{abstract}

\vspace{.5em}

\section{Introduction}

The multi-armed bandit (MAB) model serves as a canonical framework in sequential prediction.  In each round $t=1,\ldots,T$ the algorithm may query any arm $i \in [K]$, it then gets to observe a reward value $X_{t,i} \in [0,1]$, and updates its strategy for future rounds. The two primary tasks considered under this model are the \textit{best-arm identification} (BAI), also known as pure exploration, and {\it regret minimization}. 

In the identification task, we are interested in finding the best arm, defined as the arm with the highest
mean reward, using minimal number of queries $T$. On the other hand, in regret minimization, the goal is to minimize the difference between the cumulative reward achieved by the algorithm and that of the single best arm in hindsight.

While the stochastic MAB setting has been studied extensively, prior work on {\it adversarial} MAB has focused almost exclusively on regret minimization, and not addressed the BAI task. The reason for this gap is clear; the standard BAI objective is essentially futile under adversarial rewards. That is, identifying the arm with the largest {\it past} cumulative reward seems to offer no guidance about which arm will perform best in the {\it future}, precisely because of the adversarial nature of the problem.

To address this gap,   we formulate the task of \textit{lookahead BAI}, where instead of identifying the arm with the best historical performance, the goal of lookahead
BAI is to find an arm that will provide the highest expected cumulative gain
over a {\it future} time window.  The algorithm is allowed to choose the starting time, as well as the
length, of the prediction window (within pre-specified ranges), possibly depending on the observations so far.

{The lookahead BAI task is motivated by  applications of the classic BAI task in stochastic bandits \citep{bubeck2009pure, garivier2016optimal, karnin2013almost}, including online advertising, web content optimization and A/B testing, and clinical trials. In these applications the algorithm  may choose a stopping-time but must then commit to a single decision without further experimentation as additional exploration is resource-intensive. However, while this setting is well-motivated, it has not been studied in the adversarial regime. Removing unrealistic (stochastic, or other) assumptions makes the problem substantially more challenging, and to the best of our knowledge,our work provides the first positive results in adversarial best-arm identification.
}

Our first contribution is that even under \textit{no assumptions on the data}, non-trivial guarantees can be achieved with error at most $O(1/\sqrt{\log(T)})$. We then show that this bound is near-tight with a lower bound of $\Omega(1/{\log(T)})$. Next, we turn to study memory-accuracy trade-offs for lookahead-BAI, as well as for the classical regret minimization objective.

\paragraph{Memory constraints}  In recent years, the study of online learning under bounded memory constraints has attracted considerable attention. While most works consider the stochastic MAB setting \cite{agarwal2022sharp, maiti2021multi, wang2023tight}, recent work \citep{srinivas2022memory, peng2023near, peng2023online} initiated the study of memory bounds for the adversarial \textit{expert problem}, providing a tight characterization of the memory-regret tradeoff. However, this does not generalize to the bandit setting with limited feedback.

\begin{table}[!t]\captionsetup{format=plain}
\begin{center}
\begin{tabular}{||c | c c ||} 
 \hline
  &  Upper bounds  & Lower bounds  \\ [0.5ex] 
 \hline\hline
 BAI  &  
  \begin{tabular}{@{}c@{}}   $\epsilon = O(\frac{1}{\sqrt{\log{T}}})$,  { $\sigma = \tilde{O}(K)$} \\ (Theorem \ref{thm:bandit_bai})\end{tabular} 
 &      \begin{tabular}{@{}c@{}} $\epsilon = \Omega(\frac{1}{{\log{T}}})$,  { $\sigma = {\Omega}(K)$}\\ {{(Theorem \ref{thm:bandit_bai_err_lb}, Theorem \ref{thm:bandit_bai_memory_lb}) }}\end{tabular} 
  \\ 
 \hline
  BAI, Sparse case  &   
 \begin{tabular}{@{}c@{}}    $\epsilon = O(\frac{1}{\sqrt{\log{T}}})$,  { $\sigma = \tilde{O}(1)$}\\ (Theorem \ref{thm:sparse_bai})\end{tabular} 
 &  --
 \\
  \hline
  Regret   &   
 \begin{tabular}{@{}c@{}}    $R =  \tilde{O}(T^{2/3}K^{1/3})$,  { $\sigma = \tilde{O}(1)$}\\ (Theorem \ref{thm:bm_mab})\end{tabular} 
 &       \begin{tabular}{@{}c@{}} $R = \Omega(\sqrt{KT}), \sigma = \Omega(\text{poly-log}(KT))$ \\ {\footnotesize \citep{srinivas2022memory, peng2023near}}\end{tabular}
 \\
 \hline
\end{tabular} 
\end{center}
\caption{\textbf{Summary of results}. The above table presents our main results in this work, where in all cases the MAB bandit model is assumed, with $K$ arms and $T$ rounds. The bounds for the lookahead BAI task (see Definition \ref{def:lookahead BAI}) are given in the first row, for the error $\epsilon$, and $\sigma$ the total number of memory bits required. The second row also addresses the lookahead BAI task, under a general sparsity condition (see Definition \ref{def:sparse}). The last row gives regret bounds $R$, as well as memory bounds. The $\tilde{O}(\cdot)$ hides poly-logarithmic factors in both $K$ and $T$. Characterizing the lower bounds BAI in the sparse setting remains an open question.\label{tab:results_table2}
} 
\end{table}

In this work, we consider both the {lookahead BAI} and regret minimization tasks under bounded memory constraints. First, we show that while our algorithm for the lookahead BAI problem incurs a large memory cost of $\Omega(K)$, that is unavoidable in general. Then, we consider a relaxed setting in which we assume a general \textit{sparsity} condition applies. We prove that when this condition is satisfied,
significant memory improvements can be achieved. Specifically, we give an algorithm for the lookahead BAI task that requires at most $O(\text{poly-log}(KT))$ bits of memory, for a broad family of sparse bandit instances.

To understand whether this $\Omega(K)$ memory requirement is inherent to adversarial bandits or specific to lookahead prediction, we also investigate the regret minimization task. \cite{srinivas2022memory} studied the distinction between memory requirements for regret and best-arm identification in the expert setting. We provide an efficient reduction from the expert setting which yields a MAB algorithm achieving sublinear regret with significantly better memory performance than lookahead BAI. Our main results are summarized in Table \ref{tab:results_table2}.

\subsection{Prior work}\label{sec:prior-work}

\paragraph{Best-arm identification with lookahead}
{The  BAI problem, sometimes referred to as the \textit{pure exploration} problem, is well-studied under stochasticity assumptions (see e.g., \cite{garivier2016optimal, karnin2013almost, bubeck2009pure, mannor2004sample},  as well as  \cite{lattimore2020bandit}, Chapter 33).   We remark that in the BAI formulation it is often allowed that the algorithm declares on a stopping time adaptively, and must then commit to a single decision, as additional exploration can be costly. Such formulations are discussed for example in \cite{lattimore2020bandit}, in Section 33.2. Without stochastic assumptions, the problem becomes substantially more challenging, and to the best of our knowledge this work provides the first positive results for adversarial best-arm identification. }

\paragraph{Memory bounded experts}
In the adversarial setting, the problem of  Memory bounded experts was studied (and mostly resolved) by \cite{srinivas2022memory, peng2023online, peng2023near}.  The work by \cite{peng2023near} achieves a regret bound of $\tilde{O}(\sqrt{TK/\sigma})$ for $\sigma = \Omega(\text{poly-log}(KT))$. The algorithm given in their work is rather complex, and in particular does not admit a simple generalization to the more challenging bandit setting, in which only limited feedback is available.

\paragraph{Memory bounded bandits}
There are several works on bounded-memory multi-armed bandits (MAB) in the streaming, {\itshape stochastic} settings \citep{maiti2021multi, agarwal2022sharp, wang2023tight}.\footnote{We further remark that the lower bounds given in the aforementioned papers do not apply to the standard MAB setting, since in the single-pass streaming model, once an arm was removed from the pool it can never be used again.}
However, in the adversarial setting, which is the focus of this work, the only prior work on bounded-memory MAB to the best of our knowledge is by \cite{xu2021memory}. However, the performance bounds of their algorithm are inferior to those given in this work, both in terms of regret guarantees, and memory bounds. See further discussion in Section \ref{sec:bm-mab}.

\paragraph{Density prediction} Our analysis of the lookahead BAI algorithm is derived by a technique initially developed for a {\itshape density prediction} setting, introduced by \cite{drucker2013high}. In this setting, a single stream of binary bits is presented sequentially, and an algorithm is asked to predict the density of future bits within a certain window of its choice. Then, \cite{qiao2019theory} extended this setting to functions other than the average, considering more general statistics of the sequence. In contrast, the lookahead BAI task is concerned with the case that $K$ arbitrary sequences are given. Critically, the algorithm may not access the sequences, and may only query an arm and observe bandit feedback for its chosen arm.

\section{Lookahead BAI for multi-armed bandits}

 We first introduce the task of lookahead BAI, defined as follows.  We consider $K$ arms and a horizon $T$. The learner operates under a memory budget of $\sigma$ bits, and given error parameter $\epsilon > 0$.   In each round $t=1,...,T$ the algorithm may query any arm $i \in [K]$, and observe value $X_{t,i} \in [0,1]$, chosen possibly adversarially ahead of time by an oblivious adversary. 
At some (possibly randomized) stopping time $t_0\in[T]$ and with a chosen window length $w\ge 1$ such that $t_0{+}w\le T$, the learner outputs an arm $\hat{i}\in[K]$. The goal is to guarantee that the chosen arm is $\varepsilon$-optimal over the chosen window.

\begin{definition}[Lookahead BAI]\label{def:lookahead BAI}
Fix $\varepsilon>0$. An algorithm for lookahead BAI outputs 
a window size $w \in [T]$, a stopping time $t_0 \le T-w$, and an arm $\hat{i}\in[K]$, such that it achieves $\epsilon$-best expected cumulative gain over the window,
\begin{equation}
\mathbb{E}\!\left[\;\max_{i^\star\in[K]}\Bigg(\frac{1}{w}\sum_{t=t_0}^{t_0+w} X_{t,i^\star}\Bigg) - \frac{1}{w}\sum_{t=t_0}^{t_0+w} X_{t,\hat{i}} \;\right] \le \varepsilon,
\end{equation}
where the expectation is over the algorithm's randomness. 
\end{definition}

 Building on the formulation above, we give a  Lookahead BAI algorithm, given in Algorithm \ref{alg:bandit_bai}. We first prove it achieves $\epsilon$-optimality.  We then follow up in this section with two near-tight lower bounds, on the accuracy parameter, and on the memory requirements. 
\begin{algorithm}[!t]
\caption{Bandit lookahead best-arm identification} \label{alg:bandit_bai}
\vspace{.5ex}
\begin{flushleft}
  {\bf Given:}   parameters $T, K \in \mathbb{N}$, and $ \delta  > 0$\\
{\bf Output:} time step $t_0$, window size $w$ and arm $\hat{i} \in [K]$
\end{flushleft}
\begin{algorithmic}[1]
\STATE Sample $m$ uniformly from $\{\ceil*{\log(T)/2},...,\ceil*{\log(T)}\}$. Sample $b$ uniformly from $[T/2^m]$
\STATE Set $w := 2^{m}/2$ and $t_0 := (b-1)2^m + w + 1$
\STATE Initialize ${\Tilde n}_i := 0$ for all $i=1,...,K$ 
\STATE For all rounds $t=1,...,t_0-w-1$, pick any arbitrary arm, and discard its value
\FOR{$t = t_0-w, \ldots, t_0$}
\STATE Choose arm $i_t \in [K]$ uniformly at random and observe its value $X_{i_t,t}$
\STATE Update ${\Tilde n}_{i_t} \leftarrow {\Tilde n}_{i_t}  +   X_{i_t,t}$ 
\ENDFOR
\STATE Output time step $t_0$, window size $w$, and the maximizing arm:
$$\hat{i} = \arg\max_{i \in [K]} \ {\Tilde n}_i$$
\end{algorithmic}
\end{algorithm} 
  
\begin{theorem}[{\bf Bandit lookahead BAI}]\label{thm:bandit_bai}
Let $K, T \in \mathbb{N}$ such that $K = \tilde{O}({{T^{1/4}}})$.
For any bandit instance $X \in [0,1]^{K \times T}$, Algorithm \ref{alg:bandit_bai} returns $t_0$, window size $w = \Omega(\sqrt{T})$, and arm $\hat{i}$  
such that,
\begin{equation}    
  \E \left[ \max_{i^*\in[K]}  \frac{1}{w} \sum_{t=t_0}^{t_0+ w} X_{t,i^*} \ - \ \frac{1}{w} \sum_{t=t_0}^{t_0+ w} X_{t,{\hat i}}  \right] \le \frac{20}{\sqrt{{\log(T)}}},
\end{equation} 
\end{theorem}

Before giving the proof, we first give the following lemma used in the analysis of the theorem 
to estimate the values of an individual arm, in the simpler full information case. That is, assuming we can access all values of that arm sequentially, Lemma \ref{lemma:key_lemma} shows that we can find a prediction window and estimate of the average value in that window.
\begin{lemma} 
\label{lemma:key_lemma} Let $T \in \mathbb{N}$ and $X \in [0,1]^T$, and let $\underline{w}, \overline{w} \in \mathbb{N}$ with $\underline{w} < \overline{w} < \floor*{\log_2(T)}$.     Let $m$ and $b$ be sampled uniformly at random from  $\{\underline{w},...,\overline{w}\}$ and $[T/2^m]$, respectively. Set $w := 2^{m-1}$ and $t_0:= (b-1)2^m + w + 1$, and denote $y = \frac{1}{w} \sum_{t=t_0-w}^{t_0-1}  X_t$ and   $y^* = \frac{1}{w}\sum_{t=t_0}^{t_0 + w} X_t$. Then, 
    $$\E_{m,b}\left[(y - y^*)^2\right] \le 4/(\overline{w} - \underline{w}).$$
\end{lemma}
We remark that the proof of Lemma \ref{lemma:key_lemma} is derived using techniques given by \cite{drucker2013high}, in a related setting of density prediction in binary sequences. However, in Lemma \ref{lemma:key_lemma} we prove a more general result in which we can control the window size $w$ to be withing $\{2^{\underline{w}-1},...,2^{\overline{w}-1}\}$, for any choice of $\underline{w} < \overline{w} < T$, and apply it to the continuous interval $[0,1]$ and a general $T \in \mathbb{N}$.\\

\begin{proofof}{Lemma \ref{lemma:key_lemma}}
We start by first assuming that  $T=2^M$ for some $M\in\mathbb{N}$.
Construct a perfect binary tree of height $M$ whose corresponding to an element of $X$ (leftmost leaf node is $X_1$ followed by $X_2$, etc.),
and whose internal (non-leaf) nodes store averages of their descendant leaves.
Let $\mu$ denote the value stored at the root, i.e., $\mu = \sum_{t=1}^T X_t$. 
 
 For {any} depth $j \in \{0,1,...,M\}$, let $Z(j)$ denote the random variable corresponding to the value
reached by taking a (directed) random walk of length $j$ from the root, down to height $M-j$.  Thus $Z(0)=\mu$, and
$Z(M)$ is uniformly distributed over $X_1,\dots,X_T$.
 
Notice that for any fixed height $j$, the value of the random variable $Z(j)$ is equivalently determined by  uniformly sampling one element $b_j$ out of all $2^{M-j}$ height-$j$ nodes in the tree. Moreover, observe that after $M-m$ steps the random walk reaches a node whose subtree consists
of exactly $2^m$ consecutive leaves, and therefore $Z(M-m)$ equals the average over
the corresponding contiguous block of size $2^m$ in $X$. We  thus have that $2^m = 2w$ is the size of the block, $b$ is 
the index of the block, and $t_0$ is the start time. In particular, we have, 
\[
Z(M-m)=\frac{y+y^*}{2},
\] 
Moreover, taking one additional step downwards in the random walk (to height $m-1$) will yield: 
\[
Z(M-m+1)=
\begin{cases}
y, & \text{with probability } 1/2,\\
y^*, & \text{with probability } 1/2.
\end{cases}
\]
Therefore, as $Z(M{-}m) - y = -(Z(M{-}m) - y^*)$, we get that, 
\begin{equation}\label{eq:key_eq}
    \bigl(Z(M-m+1)-Z(M-m)\bigr)^2=\left(\frac{y-y^*}{2}\right)^2.
\end{equation}
Next, we would like to bound the term $(y-y^*)^2$ using the following claim. 
\begin{claim}\label{clm:orthogonality}
Let $Z(0),Z(1),\dots,Z(M)$ be defined by a random walk down a perfect binary tree whose
internal nodes store averages of the leafs, as above.
Then, for any integers $0\le L<U\le M$,
\[
\E\!\left[(Z(U)-Z(L))^2\right]
=
\E\!\left[\sum_{j=L}^{U-1}(Z(j+1)-Z(j))^2\right].
\]
\end{claim}
The proof of Claim \ref{clm:orthogonality} is deferred to the appendix.  Then, since $m$ is uniformly distributed over $\{\underline{w},\dots,\overline{w}\}$,
the depth $M-m$ is uniformly distributed over
$\{M-\overline{w},\dots,M-\underline{w}\}$.
Applying Claim \ref{clm:orthogonality} with
$L=M-\overline{w}$ and $U=M-\underline{w}$ we get, 
\[
\E\!\left[(Z(M-m+1)-Z(M-m))^2\right]
=
\frac{1}{\overline{w}-\underline{w}}
\E\!\left[(Z(M-\underline{w})-Z(M-\overline{w}))^2\right].
\]
Since all values lie in $[0,1]$, the squared difference on the right-hand side is at
most $1$, and hence
\[
\E\!\left[(Z(M-m+1)-Z(M-m))^2\right]\le
\frac{1}{\overline{w}-\underline{w}}.
\]
Combining this with the previous identity given in \eqref{eq:key_eq} yields,
\[
\E[(y-y^*)^2]\le \frac{4}{\overline{w}-\underline{w}}.
\] 
Finally, we consider the case that $T$ is not a power of $2$. In this case, we take $T'  = 2^{\floor*{\log_2(T)}}$, and apply the above argument $T'$ rather than $T$, and over $X$ restricted to $T'$. 
\end{proofof}

\noindent We are now ready to give the proof of Theorem \ref{thm:bandit_bai}, proving the accuracy guarantees of Algorithm \ref{alg:bandit_bai}. \\

\begin{proofof}{Theorem \ref{thm:bandit_bai}}
 For each $i = 1,...,K$ define $y_i = \frac{1}{w}\sum_{t=t_0-w}^{t_0-1}  X_{t,i}$, where $t_0$ and $w$ are set as in Algorithm \ref{alg:bandit_bai}. This is the true sum of rewards for each arm on the window of size $w$, following $t_0$. This quantity cannot be accessed by the algorithm, which instead produces an estimator ${\tilde y}_i$. We first show that the error for prediction based on the true sums $y_i$ is bounded.  Let $z_i = \frac{1}{w}\sum_{t=t_0}^{t_0 + w} X_{t,i}$.  Let $M := \floor*{\log(T)/2}$. 
 By Lemma \ref{lemma:key_lemma} we get that  $\E_{m,b}\left[(y_i - z_i)^2\right] \le 4/M$, 
and so by concavity of $\sqrt{\cdot}$ and Jensen's inequality, we have,
$$
\E_{m,b}\left[|y_i - z_i|\right] \le \frac{2}{\sqrt{M}} =: \epsilon_1.
$$ 

Next, since we do not have access to the $y_i$'s but only the count estimates ${\tilde n}_i$. Define $$\tilde{y}_i = \frac{1}{w}\sum_{t=t_0-w}^{t_0-1}X_{t,i}\cdot \mathbf{1}[i_t = i] \cdot K = \tilde{n}_i \cdot \frac{K}{w}.$$  Notice that 
when taking expectation over the random choices of $i_{t_0-w},...,i_{t_0-1}$ (and for {\it fixed} $t_0$ and $w$), we have that for all $i = 1,...,K$,
$$
\E[\tilde{y}_i] = y_i.
$$
Let $\epsilon_2 > 0$. By Hoeffding's inequality, for any $i \in [K]$ with probability $1-\frac{\epsilon_1}{2}$ it holds  that, 
$$
\Pr[|{\tilde y}_i - y_i| \ge \epsilon_2] = \Pr[|{\tilde n}_i \cdot K - y_i \cdot w| \ge \epsilon_2 \cdot w] \le 2e^{-2w\epsilon_2^2/K^2} \le \epsilon_1,
$$
where the randomness is only over the choices of $i_{t_0-w},...,i_{t_0-1}$ as in Line 6 of Algorithm \ref{alg:bandit_bai}, and by setting $\epsilon_2 = \sqrt{\frac{2K^2\ln(2/\epsilon_1)}{\sqrt{T}}}$, we have that the above is bounded by $2e^{-\ln(2/\epsilon_1)} < {\epsilon_1}$, 
since $w > \sqrt{T}/4$. Thus, all $i \in [K]$ we have that $\E |{\tilde y}_i - y_i| \le (1-\epsilon_1) \epsilon_2 + \epsilon_1 \le \epsilon_1 + \epsilon_2$. Moreover, by triangle inequality  we have,
\begin{equation}\label{eq:triangle}
    \E |{\tilde y}_i - z_i| \le \E |{\tilde y}_i - y_i| + \E |y_i - z_i| \le 2\epsilon_1+ \epsilon_2 =:  \epsilon.
\end{equation}
Therefore, we have that $\hat{i}$ chosen in Algorithm \ref{alg:bandit_bai} satisfies,
$$
\E [z_{\hat i} ]\ge \E [{\tilde y}_{\hat i}] - \epsilon \ge  
\E [{\tilde y}_{i^*}] - \epsilon \ge 
\E  [z_{i^*}] - 2\epsilon,
$$
for $i^* = \arg\max_{i \in [K]} z_i$, 
where the first and last inequality follow by Eq. \eqref{eq:triangle}, and the second inequality follows by the choice of $\hat{i}$ in Algorithm \ref{alg:bandit_bai} to be the maximizer of the $\tilde{n}_i$'s, which is thus also the maximizer of the $\tilde{y}_i$'s. 
We have that for any integer $T \ge 2$, and $K = O(\frac{{T^{1/4}}}{\log(T)})$,
$$
\epsilon \le \frac{8}{\sqrt{\log(T)}} + \sqrt{\frac{2K^2\ln(\sqrt{\log(T)})}{\sqrt{T}}} \le \frac{10}{\sqrt{\log(T)}}.
$$
Thus, by also plugging in the definitions of $z_{\hat{i}}, z_{i^*}$, this concludes the proof.
\end{proofof}

\subsection{Lower bounds}

This section is concerned with characterizing two lower bounds for the lookahead BAI setting. One result focuses on studying the overall accuracy of lookahead BAI algorithms, the other on identifying the minimum amount of memory required to guarantee any non-trivial error.

We start by showing that in the general non-memory-bounded setting, the performance of Algorithm \ref{alg:bandit_bai} guaranteed by Theorem \ref{thm:bandit_bai} is potentially improvable only by a square root, but still incurs a logarithmic dependency with respect to the sequence length. 

\begin{theorem}[\textbf{Lower bound for lookahead BAI}]\label{thm:bandit_bai_err_lb}
    Let $K=2$ and $T = 2^M$ for $M \in \mathbb{N}$. Then, for any algorithm for the lookahead BAI task, there exists an instance $X \in [0,1]^{K \times T}$ for which the algorithm must incur an expected error of at least $1/{(8\log(T))}$.
\end{theorem}
\paragraph{Proof sketch} The full proof of Theorem \ref{thm:bandit_bai_err_lb} is deferred to \Cref{app:missing_proofs}, however here we will describe the main ingredients of the proof. We will prove a stronger claim, by considering any algorithm that may {\it fully observe} the values of {\it both} arms $i \in [K]$, and is only required to pick a stopping time $t_0$ after which, it returns an arm $\hat{i}\in [K]$ and window size $w$ such that,
\begin{equation}  \label{eq:lb_alg_def_1}
  \E \left[ \max_{i^*\in[K]}  \frac{1}{w} \sum_{t=t_0}^{t_0+ w} X_{t,i^*} \ - \ \frac{1}{w} \sum_{t=t_0}^{t_0+ w} X_{t,{\hat i}}  \right] \le \epsilon,
\end{equation}
where the expectation is over the randomness of the algorithm.  We will show that any such algorithm must incur expected error $\epsilon \ge 1/(8{\log(T)})$.

We now describe a construction of a random instance $X^{2 \times T} \in [0,1]$. Consider a perfect binary tree $\mathcal{T}$ of height $M$, with $2^M$ leaf nodes. We consider functions $f$ which assign a value $f(v) \in [0,1]$ to each node $v \in \mathcal{T}$.  We define a distribution $\mathcal D$ over assignment functions $f$ in a  top–down way as follows. For the root node $v$, its value is always $f(v) = 1/2$, and its sign is $S_v = +1$. 
    For each non-root node $v$ at depth $d\in\{1,\dots,M\}$ we assign it a value $f(v) \in [0,1]$ and sign $S_v \in\{\pm1\}$ at random, based on its parent node sign. Namely, if $p$ is the parent node of $v$, then,
    $$
    S_v =\begin{cases}
			+S_p, & \text{w.p. $\alpha_d$,}\\
            -S_p, & \text{w.p. $1-\alpha_d$,}
		 \end{cases}
    $$
    where $\alpha_d := \frac{1}{2}\left(1 + \sqrt{1-\frac{1}{d}}\right)$. 
Its value is then $f(v) = \frac{1}{2}\left(1 + S_v\sqrt{d/M}\right)$. The randomized instance $X^{2 \times T} \in [0,1]$ is given by the $T=2^M$ leaf node values of two assignments denoted $f_1,f_2$ sampled independently from $\mathcal D$.

We now give two simple claims (whose proofs are deferred to \Cref{app:missing_proofs}), where Claim \ref{clm:lb_main_clm}  below will be useful for the remainder of the proof of the theorem.
\begin{claim}\label{clm:simple_clm1}
    For any non-root node $v$, the marginal distribution of its sign is $\Pr_{\mathcal{D}}[S_v = +1] = 1/2$.
\end{claim}

\begin{claim}\label{clm:lb_main_clm}
    Fix any parent node $v$ at depth $d-1$ for some $d\ge 1$, denote its two children $v_L$ and $v_R$, and 
let $h: \{\pm 1\}^2 \rightarrow \{1,2\}$. Then, 
$$
\Pr_{f_1,f_2 \sim \mathcal{D}} \left[ S_{v_R}^{(1)} \neq S_{v_R}^{(2)} \ \land \ h(S_{v}^{(1)}, S_{v}^{(2)}) \notin \arg\max_{i} S_{v_R}^{(i)}  \right] \ge \frac{1}{8d}.
$$
\end{claim}

Let $A$ be any algorithm for the simplified task described above \Cref{eq:lb_alg_def_1}. We first consider any {\it deterministic} algorithm $A$ for the above task, where the observed values are randomized. That is, the algorithm $A$  chooses $t_0, w$, and arm ${\hat i}$ deterministically as a function of the (randomized) values it observes.
Notice that for any choice of $t_0$ and $w$, this determines a parent node $v$ and its two children $v_L$ and $v_R$, such that the algorithm observes all the values $x_1^{(1)},...,x_{t_0}^{(1)}$ and $x_1^{(2)},...,x_{t_0}^{(2)}$, and in particular can deduce both $f_1(v_L)$ and $f_2(v_L)$, and then needs to return $i \in \{1,2\}$ such that $i \in \arg\max_{i^*} f_{i^*}(v_R)$. 
Furthermore, observe that since for any $i \in [K]$, we have that by definition,
\begin{equation}\label{eq:lb_gap}
    |f_{1}(v_R) - f_{2}(v_R)| = \begin{cases}
     \sqrt{d/M}   & \text{ if }S_{v_R}^{(1)} \neq S_{v_R}^{(2)},\vspace{1em} \\  
    0 & \text{ otherwise. }
\end{cases}
\end{equation}

Thus, whenever the two signs of the right-hand-side node $v_R$ differ and the algorithm makes an incorrect guess, it will incur an error of $\ge   \sqrt{d/M}$. Therefore, it remains to lower bound the expected probability of error, as we show in \Cref{app:missing_proofs}, by utilizing Claim \ref{clm:lb_main_clm}.

\paragraph{Memory lower bound} We now focus our attention on the memory constraints of the lookahead BAI algorithm. We prove that any algorithm for the lookahead BAI task requires a $\Omega(K)$-bit memory. The proof follows by a reduction to the well-known   two-party Set-Disjointness  problem in communication complexity \citep{kalyanasundaram1992probabilistic, bar2004information}. We remark that our proof is similar to the remark mentioned in \cite[Sec. 1.4]{srinivas2022memory}, yet requires a different treatment compared to the standard BAI task, as in the lookahead case the algorithm may randomly choose the stopping time and window.

\begin{theorem}[{\bf Memory lower bound}]\label{thm:bandit_bai_memory_lb}
Let $K,T\in\mathbb{N}$ and fix a constant $c>1$ with $T|c$. Let $w_0 =T/c$ and 
consider any algorithm $\mathcal A$ which 
solves the lookahead best-arm identification task using any $w > w_0$ and a randomized $t_0$ declared ahead of the run, and outputs the $\epsilon$-best arm  with constant probability and, for $\epsilon \in (1/10, 1/5)$. Then $\mathcal A$ must use $\Omega(K)$ bits of memory.
\end{theorem}

\begin{remark}
    We remark that for the accuracy parameter $\epsilon$ in Theorem \ref{thm:bandit_bai_memory_lb}, it is possible to get an upper bound for a large window $w > T/2$, and with randomized $t_0$ declared ahead of the run, matching Theorem \ref{thm:bandit_bai_memory_lb} and thus obtaining a tight bound on the memory requirement. This holds by a slight modification of Algorithm \ref{alg:bandit_bai}, where $m$ is chosen uniformly at random from $\{\ceil*{\log(T)-100},...,\ceil*{\log(T)}\}$. Due to Lemma \ref{lemma:key_lemma}, we may allow for a large window when the error parameter is kept large. 
\end{remark}

\section{Lookahead BAI with bounded memory for sparse bandits}

In this section we consider a {\it local sparsity} condition of bandit instances to yield improved   memory bounds. 
Here we consider for simplicity the case that each arm has a reward values in $\{0,1\}$, although this assumption can be relaxed as described later in this section.

\begin{definition}[Locally sparse bandits]\label{def:sparse}
    Let $X \in \{0,1\}^{K\times T}$ denote a bandit instance.   Denote by $\Bar{n} \in [T]^K$ the sum of rows of $X$, i.e., $\Bar{n}(i) = n_i =  \sum_{t} X_{i,t}$, and w.l.o.g assume the rows of $X$ are ordered such that $n_1 \ge n_2 \ge ... \ge n_K$.      We say that $X$ is a $\phi$-\emph{sparse} bandit instance  if $\frac{\lVert \Bar{n} \rVert_2^2}{n_1^2} \le \phi$.   We say that $X$ is a \emph{locally} $\phi$-\emph{sparse} bandit instance with window size $w \in [T]$, if for every contiguous set of indices $I \subset [T]$ of size $w$, then $Y|_{I}$ the restriction of $X$ to the columns in $I$, is  $\phi$-\emph{sparse}. 
\end{definition}

The following claim demonstrates that the above condition is rather general, and provides a concrete example of a broad family of instances which satisfy Definition \ref{def:sparse}, where its proof is deferred to \Cref{app:missing_proofs}.
\begin{claim}[Polarized bandits]\label{example:sparse_l2}
    Let $K, T \in \mathbb{N}$ such that $K^2 \le T$, and window size $w \ge \sqrt{T}$. Let $X \in \{0,1\}^{K\times T}$, and let $\Bar{n} \in [T]^K$ be the sum of its rows.
Assume  there are $r \ge 1$  elements $i \in [K]$ with  $n_i \ge T - w/2$, and all other $K-r$ elements $j$ have  $n_j =  O(\sqrt{w})$. Then, $X$ is locally  $\phi$-{sparse} for $w$, with $\phi = O(r)$.
\end{claim}

In what following we define a streaming setting and a task in this model called \emph{ApproxTop}, and give a classical result by \cite{charikar2004finding} which states that the well-known CountSketch algorithm solves the \emph{ApproxTop} task. Then, we apply CountSketch within Algorithm \ref{alg:sparse_bandit_bai}, and prove in Theorem \ref{thm:sparse_bai} that by a reduction from the bandit setting to \emph{ApproxTop}, we indeed get improved memory bounds, compared to those given in the previous section. 

\begin{definition}\label{def:approxtop}
    Let $S = q_1,q_2,...,q_n$ be a data stream, where each $q_i \in O = \{o_1,...,o_k\}$. Let object $o_i$ occur $n_i$ times in $S$. Order the $o_i$ so that $n_1 \ge n_2 \ge \ldots \ge n_k$, and let $\Bar{n} = (n_1,n_2,...,n_k)$. We say that a $1$-pass algorithm over the stream solves the \emph{ApproxTop}$(S,\epsilon, \delta)$ problem if it returns element $i$ such that $n_i \ge (1-\epsilon)n_1$ with probability at least $1-\delta$.
\end{definition}

\begin{theorem}[\cite{charikar2004finding}, Theorem 1]\label{thm:countsketch}
    The \emph{CountSketch} algorithm solves the \\
    \emph{ApproxTop}$(S,\epsilon, \delta)$ problem using space $O({\phi_S} \cdot \frac{1}{\epsilon^2}\log \frac{n}{\delta})$, where $\phi_S = \frac{\lVert \Bar{n} \rVert_2^2}{n_1^2}$. 
\end{theorem}

\begin{algorithm}[t]
\caption{Sparse-bandit lookahead best-arm identification} \label{alg:sparse_bandit_bai}
\vspace{.5ex}
\begin{flushleft}
  {\bf Given:}   parameters $T, K \in \mathbb{N}$, and $\phi$,  and $\delta', \epsilon' > 0$  \\
{\bf Output:} time step $t_0$, window size $w$ and arm $\hat{i} \in [K]$
\end{flushleft}
\begin{algorithmic}[1]
\STATE Sample $m$ uniformly from $\{\floor*{\log(T)/2},...,\floor*{\log(T)}\}$. Sample $b$ uniformly from $[T/2^m]$
\STATE Initialize ${\Tilde y}_i := 0$ for all $i=1,...,K$.
\STATE Set $w := 2^{m}/2$ and $t_0 := (b-1)2^m + w + 1$
\STATE \textcolor{Blue}{Initialize CountSketch to be run over $k = K$ items with $\phi$ and $\delta', \epsilon'$ } 
\FOR{$t = t_0-w, \ldots, t_0$}
\STATE Choose arm $i_t \in [K]$ uniformly at random and observe its value $X_{i_t,t}$
\STATE \textcolor{Blue}{Update CountSketch with item $i_t$ if value $X_{i_t,t} =1$ }
\ENDFOR
\STATE \textcolor{Blue}{Output time step $t_0$, window size $w$, and the arm $\hat{i} \in [K]$ returned by CountSketch}
\end{algorithmic}
\end{algorithm} 
\vspace{-.3em}

 We are now ready to give our main result for the lookahead BAI, under the assumption that the instance is locally $\phi$-sparse. The proof of Theorem \ref{thm:sparse_bai} is deferred to \Cref{app:missing_proofs}.
\begin{theorem}[{\bf Sparse bandit lookahead BAI}]\label{thm:sparse_bai}
Let $K, T \in \mathbb{N}$ such that  $K = \tilde{O}({{T^{1/4}}})$\footnote{$\tilde{O}(\cdot)$ hides poly-logarithmic factors in $T$.}, and let $\delta >0$. Set $\epsilon' = 2/\sqrt{\log(T)}$, $\delta' = \delta\epsilon'$.
For any bandit instance $X \in [0,1]^{K \times T}$ that is locally $\phi$-sparse for all $w \ge \sqrt{T}$, with probability at least $1-\delta$ we have that Algorithm \ref{alg:sparse_bandit_bai} uses  at most $\sigma = \tilde{O}(\phi)$ bits of memory, and
returns $t_0$, size $w = \Omega(\sqrt{T})$ and arm $\hat{i}$  
such that \begin{equation}    
  \E \left[ \max_{i^*\in[K]}  \frac{1}{w} \sum_{t=t_0}^{t_0+ w} X_{t,i^*} \ - \ \frac{1}{w} \sum_{t=t_0}^{t_0+ w} X_{t,{\hat i}}  \right] \le \frac{10}{\sqrt{{\log(T)}}}.
\end{equation}
\end{theorem}

\begin{remark}
    Consider the general instance given in Claim \ref{example:sparse_l2}, for $r = O(\text{poly-log}(KT))$ such that most items have $O(\sqrt{T})$ total reward, and  our task is to return any of the $r$ {\it heavy} items with $\Omega(T)$ reward. Then, Theorem \ref{thm:sparse_bai} implies that Algorithm \ref{alg:sparse_bandit_bai}
    achieves this for any $T$ using only poly-logarithmic memory. 
\end{remark}
 
\vspace{-1.2em}

\paragraph{Extension to real-valued rewards $X \in [0,1]^{K \times T}$} We remark that although the above algorithm and analysis are given for the case that the rewards have binary values, they can easily be adapted to the continuous case as follows. 
Although the analysis of CountSketch given in 
\cite{charikar2004finding}
primarily focuses on frequency (unweighted) counts, the extension to the \textit{weighted} case is a direct and well-known (as described e.g., in 
\cite{cormode2008finding}), as the linearity of the sketch allows for such updates. The adjustment  required is simply that the when updating entries in the sketch, the value of the hash function is multiplied by the corresponding weight $X_{i, t}$.

\begin{algorithm}[t]
\begin{algorithmic}[1]
\STATE \textbf{input:} oracle access to $\A$ a
$(\sigma,s)$-bounded memory online learner for the expert setting, \\
\hspace{1cm} \ integer $Q$, for the number of blocks $B_1,...,B_Q$ of size $T/Q$ each. 
\FOR{$\tau = 1, \ldots, Q$}
\STATE Obtain $p_\tau$ from $\A$, and denote $J_\tau = \{j_{\tau 1},...,j_{\tau s}\}$ the arms in the support of $p_\tau$. 
\STATE Pick $s$ explorations rounds $E_\tau = \{t_{j_{\tau1}},...,t_{j_{\tau s}}\}
\subset B_\tau$ uniformly at random. 
\FOR{$t \in B_\tau$}
\STATE If $t = t_j \in E_\tau$ for some $j\in J_\tau$ then play arm $i_t = j$. Otherwise, play arm $i_t \sim p_\tau$.
\STATE Observe loss value $\ell_{t}(i_t)$
\ENDFOR
\STATE Construct loss $\hat{c}_\tau$ such that for each $j\in J_\tau$, $\hat{c}_\tau(j) := \ell_{t_j}(j)$ and zero otherwise. 
\STATE Obtain the next $p_{\tau+1}$ from $\A$ after observing the loss $\hat c_\tau$. 
\ENDFOR
\end{algorithmic}
\caption{bounded-memory MAB \label{alg:bm_mab}
} 
\end{algorithm}

\section{Regret minimization for bandits with bounded memory}\label{sec:bm-mab}

We have shown that lookahead BAI requires $\Omega(K)$ memory to achieve non-trivial accuracy in the worst-case. We now investigate whether this memory requirement is inherent to adversarial bandits, or specific to the prediction task. To this end, we study the classical regret minimization objective under memory constraints, beginning with the notion of a bounded-memory online learner.

\begin{definition}[Bounded-memory online learner] \label{def:bm-hedge}
Let $K$ be the number of experts, and $T$ be the horizon length.
An online learning algorithm $\A$ is a $(\sigma,s)$-\textbf{bounded-memory online learner}  
if for any sequence of losses $\ell_t \in [0,1]^K$, at every iteration $t \in [T]$ the algorithm picks a distribution $p_t \in \Delta_K$ such that 
the support of $p_t$ has cardinality at most $\|p_t\|_0 \leq s$ and the total memory used throughout the whole process is $\sigma$ bits.
We denote the additive regret of $\A$ as
\[
  \sum_{t=1}^T p_t \cdot \ell_t  - \min_{i^* \in [K]}  \sum_{t=1}^T  \ell_t(i^*) \le R_{\A}(T),
\]
with $R_{\A}: \mathbb{N} \rightarrow \R_+$ 
 a non-decreasing, sublinear function of~$T$. 
\end{definition}
Online learning with memory constraints have been studied by \cite{srinivas2022memory}, where they showed that for a wide range of sequences, including purely adversarial ones, any $(\sigma,s)$-bounded-memory learner achieves regret not smaller than $R_\A(T) = \Omega(\sqrt{KT/\sigma})$. While in the original paper the authors provided algorithms capable of achieving such behavior in the case of randomly sampled sequences, \cite{peng2023near} proposed an algorithm capable of attaining the lower bound above when memory was of order $\sigma = O(\text{poly-log}(KT))$. \\

By leveraging these ideas, we propose Algorithm \ref{alg:bm_mab} to tackle the bandit setting. The algorithm operates by dividing the sequence of length $T$ in $Q$ subsequent rounds $B_\tau$, for $\tau=1,\dots,Q$, with the aim of running a $(\sigma,s)$-memory-bounded online learner $\A$ over a sequence of losses $\hat c_\tau$ purposely built for each round $\tau$. More precisely, we make use of the sparsity constraint over the $p_\tau$ probability vectors produced by $\A$ to make sure that only a selected pool of arms is observed for each block $B_\tau$. At each round $\tau$, the loss $\hat c_\tau$ is obtained by a combination of exploitatory sampling according to $p_\tau$ and exploration rounds sampled uniformly at random, with the goal of informing $\A$ of the expected average loss incurred over the restricted pool of arms throughout the rounds in $B_\tau$. At the next round $\tau+1$ a new pool of arms is produced by $\A$ as support of $p_{\tau+1}$, while still guaranteeing both $\sigma$ and $s$ to be constrained. 

In the following, we show that Algorithm \ref{alg:bm_mab} is able to achieve non-linear regret while maintaining a poly-logarithmic  memory footprint $\sigma$ with respect to both $K$ and $T$.

\begin{theorem}\label{thm:bm_mab}
For sufficiently large $K$ and $T$, there exists an algorithm that achieves regret at most $\tilde O(T^{2/3}K^{1/3})$ in the bandit setting with probability at least $1-2/(KT)^{O(1)}$,
using at most $\sigma=O(\text{poly-log}(KT))$ bits of memory.  
\end{theorem}
\vspace{-1.5em}

\paragraph{Proof sketch}
The proof of Theorem \ref{thm:bm_mab} is deferred to \Cref{app:missing_proofs}. However, here we give a high-level proof sketch. We start by claiming that there exists a $(\sigma,s)$-memory-bounded learner $\A$ that achieves sublinear regret with a small memory footprint, which in turn will enable us to characterize the behavior of Algorithm \ref{alg:bm_mab}. In particular, for sufficiently large $K$ and $Q$, Algorithm $8$ in \cite{peng2023near} was shown to achieve regret $R_\A(Q) = \tilde O(\sqrt{QK/\sigma})$ with probability at least $1-1/(KT)^{\omega(1)}$. Then, we apply a reduction strategy from the bandit feedback to the full-information setting, following that given  by \cite{blum2007learning} to control the regret of Algorithm \ref{alg:bm_mab}, as given in Algorithm \ref{alg:bm_mab}.

Theorem \ref{thm:bm_mab} shows that it is possible to achieve sublinear regret even when constraining the memory available to the online algorithm significantly. Evidently, there is a significant gap in terms of the regret of $\tilde O(T^{2/3}K^{1/3})$ that Algorithm \ref{alg:bm_mab} is capable of achieving according to Theorem \ref{thm:bm_mab} in contrast to the $\tilde O(\sqrt{KT})$ reported by \cite[Thm. 1.1]{peng2023near} in the expert setting. We note however that, to date, the memory-bounded bandit setting proves more challenging to study than its full-information counterpart, with \cite{xu2021memory} proposing an algorithm that guarantees regret at most $O(T^{3/4}K^{1/4})$ while requiring $O(\sqrt{K})$-bits of memory.

Therefore, Theorem \ref{thm:bm_mab} currently offer the tightest behavior for a memory-bounded algorithm in the bandit setting, both in terms of regret upper bound and memory lower bound. This observation leaves open the question of whether an algorithm requiring only $O(\text{poly-log}(KT))$ memory exists such that it is capable of achieving $O(\sqrt{KT})$ regret also in the bandit setting or if $\tilde O(T^{2/3}K^{1/3})$ is optimal, highlighting a gap between expert and bandit settings, similar to what already observed in applications such as switching with cost \cite{dekel2014bandits}.

~\newline\noindent{\bfseries Memory constraints in regret vs. best-arm identification settings.} The results from Theorem \ref{thm:bm_mab} and Theorem \ref{thm:bandit_bai_memory_lb} establish a concrete separation between regret minimization and best-arm identification under memory constraints in the bandit setting. While \cite{srinivas2022memory} observed this distinction in the expert setting, our work demonstrates it persists in the bandit feedback model with regret minimization achieves sublinear regret with $O(\text{poly-log}(KT))$ memory whereas lookahead BAI requires $\Omega(K)$ memory.

While the memory footprint of the lookahead BAI task is large in the worst-case, we proved that under a broad sparsity condition, a significant improvement in the memory cost can be achieved, while maintaining the same accuracy level. It remains as an open question whether a similar lower bound can be proved for the sparse case. 
Together, these results make progress towards characterizing the trade-offs between achievable performance and memory bounds for adversarial bandits.

\bibliography{bib}
 
\newpage

\appendix

\crefalias{section}{appendix} %

\section{Deferred proofs}\label{app:missing_proofs}

In this section we provide proofs deferred from the main body of the paper.\\

\begin{proofof}{Claim \ref{clm:orthogonality}} Let $c_1,\dots,c_M\in\{0,1\}$ denote the random left/right choices of the walk. For any $j\in\{0,\dots,M\}$, write $\mu_{c_1,\dots,c_j}$ for the value stored at the node
reached after $j$ steps, so that
\[
Z(j)=\mu_{c_1,\dots,c_j}.
\]

Fix $j\in\{0,\dots,M-1\}$. Conditioning on the first $j$ steps of the walk, the next step
chooses one of the two children uniformly. Since each internal node stores the average
of its two children, we have
\begin{equation}\label{eq:cond-exp-zero}
\E\!\left[ Z(j+1)\mid c_1,\dots,c_j \right]
=
\frac{1}{2}\bigl(\mu_{c_1,\dots,c_j,0}+\mu_{c_1,\dots,c_j,1}\bigr)
=
\mu_{c_1,\dots,c_j}
=
Z(j),
\end{equation}
and therefore
\[
\E\!\left[ Z(j+1)-Z(j)\mid c_1,\dots,c_j \right]=0.
\]

Thus, for any $0 \le s < j < M$ and for any fixed outcome of the bits $c_2,...,c_j$, which determine the values of $Z(s), Z(s+1)$, and $Z(j)$, we have,
\begin{align}\label{drk:eq1}
    \E_{c_2,...,c_{j+1}} \Big[ &(Z(s{+}1) - Z(s))(Z(j{+}1) - Z(j))  \Big]  \\
    &= \E_{c_2,...,c_{j}} \left[ \E_{c_{j{+}1}}\left[ (Z(s{+}1) - Z(s))(Z(j{+}1) - Z(j)) \Big| c_2,...,c_{j} \right] \right] = 0,
\end{align} 
where the last equality follows from \eqref{eq:cond-exp-zero}.
Thus, for any $s<j$, the cross terms between increments at different depths vanish. Next, observe that
\[
Z(U)-Z(L)=\sum_{r=L}^{U-1} \bigl(Z(r+1)-Z(r)\bigr).
\]
Expanding the square and taking expectation gives
\begin{align*}
\E\!\left[(Z(U)-Z(L))^2\right]
&=
\E\!\Bigg[
\sum_{r=L}^{U-1} (Z(r+1)-Z(r))^2\\
& \qquad \qquad  \qquad   \qquad +
2\sum_{L\le s<j\le U-1}
(Z(s+1)-Z(s))(Z(j+1)-Z(j))
\Bigg] \\
&=
\E\!\left[
\sum_{r=L}^{U-1} (Z(r+1)-Z(r))^2
\right],
\end{align*}
where the second equality uses \eqref{drk:eq1} for the vanishing of all cross terms.
\end{proofof}

\begin{proofof}{Claim \ref{example:sparse_l2}}
We assume w.l.o.g that $n_1 \ge ... \ge n_K$, and so for all $i \le r$, $n_i \ge T - w/2$ by assumption. Observe that for every $w$-length window $I$, it must hold that $n_i(I) \ge w - w/2 = w/2$, where $n_i(I) = \sum_{t \in I} X_{i,t}$. Moreover, we have,
$$
\phi = \max_{w\text{-length window }}\frac{\sum_{i=1}^K n_i(I)^2}{\max_i n_i(I)^2} \le \frac{r \cdot w^2 + (K-r)O(\sqrt{w})}{(w/2)^2} \le 4r + \frac{4(K-r)}{T^{3/4}} = O(r).
$$
 \end{proofof}

\begin{proofof}{Theorem \ref{thm:sparse_bai}}
Fix a given run of Algorithm \ref{alg:sparse_bandit_bai} and corresponding instantiation of $t_0, w$, and $i_{t_0 - w},...,i_{t_0-1}$, and set $I = \{t_0-w,...,t_0-1\}$.
For each $i = 1,...,K$ define $y_i = \sum_{t \in I}  X_{t,i}$, the true sum of rewards for each arm over the window $I$. This quantity cannot be accessed by the algorithm.
We consider the following estimator,
$$
{\tilde y}_i = \frac{1}{w}\sum_{t \in I}  X_{t,i} \cdot \mathbf{1}[i_t = i] \cdot K.
$$
We also denote by  $\tilde{n}_i = \sum_{t \in I}  X_{t,i} \cdot \mathbf{1}[i_t = i]$. 
Importantly, the algorithm cannot keep track of all of the $\tilde{y}_i$ or  $\tilde{n}_i$ values, as this will incur a large memory cost. Instead, the algorithm's use of CountSketch will give a randomized estimation of the maximizer among the $\tilde{y}_i$'s, as we describe later in the proof. 

The first part of the proof follows similarly to that of Theorem \ref{thm:bandit_bai}. Specifically, we first show that the error for prediction based on the true sums $y_i$ is bounded.  Let $z_i = \frac{1}{w}\sum_{t=t_0}^{t_0 + w} X_{t,i}$.  Let $M := \floor*{\log(T)/2}$. 
 By Lemma \ref{lemma:key_lemma} we get that  $\E_{m,b}\left[(y_i - z_i)^2\right] \le 4/M$, 
and so by concavity of $\sqrt{\cdot}$ and Jensen's inequality, we have,
$$
\E_{m,b}\left[|y_i - z_i|\right] \le \frac{2}{\sqrt{M}} =: \epsilon_1.
$$

  Next, notice that when taking expectation 
over the random choices of $i_{t_0-w},...,i_{t_0-1}$ (and for {\it fixed} $t_0$ and $w$), we have that for all $i = 1,...,K$,
$$
\E[\tilde{y}_i] = y_i.
$$
Let $\epsilon_2 > 0$. By Hoeffding's inequality, for any $i \in [K]$ with probability $1-\frac{\epsilon_1}{2}$ it holds  that, 
$$
\Pr[|{\tilde y}_i - y_i| \ge \epsilon_2] = \Pr[|{\tilde n}_i \cdot K - y_i \cdot w| \ge \epsilon_2 \cdot w] \le 2e^{-2w\epsilon_2^2/K^2} \le \epsilon_1,
$$
where the randomness is only over the choices of $i_{t_0-w},...,i_{t_0-1}$ as in Line 6 of Algorithm \ref{alg:bandit_bai}, and by setting $\epsilon_2 = \sqrt{\frac{2K^2\ln(2/\epsilon_1)}{\sqrt{T}}}$, we have that the above is bounded by $2e^{-\ln(2/\epsilon_1)} < {\epsilon_1}$, 
since $w > \sqrt{T}/4$. Thus, all $i \in [K]$ we have that $\E |{\tilde y}_i - y_i| \le (1-\epsilon_1) \epsilon_2 + \epsilon_1 \le \epsilon_1 + \epsilon_2$. Moreover, by the triangle inequality  we have,
\begin{equation}\label{eq:triangle2}
    \E |{\tilde y}_i - z_i| \le \E |{\tilde y}_i - y_i| + \E |y_i - z_i| \le 2\epsilon_1+ \epsilon_2.
\end{equation}

Lastly, we apply the guarantee of the CountSketch algorithm as follows. The elements fed into CountSketch in Line 7 of Algorithm \ref{alg:sparse_bandit_bai} can be viewed as a data stream of length $n=O(w)$ with $k=K$ objects, as defined in Definition \ref{def:approxtop}. Then, observe that the count vector of the stream $\Bar{n}$ is such that each $n_i = \tilde{n}_i$, where $\tilde{n}_i$ was defined above, and satisfies $\tilde{y}_i = \tilde{n}_i \cdot \frac{K}{w}$. 
The CountSketch algorithm is also fed with $\phi$ the local-sparsity parameter of $X$, and parameters $\delta' := \delta\epsilon_1/2$ and $\epsilon' := \frac{\epsilon_1}{2}$.
Therefore, by Theorem \ref{thm:countsketch} we have that item $\hat{i}$ returned by the algorithm satisfies with probability at least $1-\delta' \ge 1-\delta$, that $\tilde{n}_{\hat i} \ge (1-\epsilon')\max_{i}\tilde{n}_{i}$. In addition, notice that for any $i \in [K]$, by fixing all randomness of the algorithm except the choices of the $i_t$'s, we have,
\begin{equation}\label{eq:eq11}
\E_{i_t, t \in I}[\tilde{n}_i] \le w/K.     
\end{equation}
Then, by taking expectation of $\tilde{n}_i$'s with respect to the randomness of CountSketch, we get that for all $i=1,...,K$ it holds that,
$$
\E[\tilde{n}_{\hat i}] \ge (1-\delta')(1-\epsilon')\E[\max_{i}\tilde{n}_{i}],
$$
and by also taking into account the randomness of the $i_t$'s and applying Eq. \eqref{eq:eq11} we get, 
$$
\E[\tilde{n}_{\hat i}] \ge \E[\max_{i}\tilde{n}_{i}] - (\frac{\delta\epsilon_1}{2} + \epsilon')\frac{w}{K} \ge \E[\max_{i}\tilde{n}_{i}] -  \epsilon_1\cdot \frac{w}{K}.
$$
 Thus, we have,
\begin{equation}\label{eq:cs_final_bound}
    \E[\tilde{y}_{\hat i}]  \ge \E[\max_{i}\tilde{y}_{i}] - \epsilon_1.
\end{equation}

Therefore,  we have that $\hat{i}$ chosen in Algorithm \ref{alg:sparse_bandit_bai} satisfies,
$$
\E [z_{\hat i} ]\ge \E [{\tilde y}_{\hat i}] - 2\epsilon_1 - \epsilon_2 \ge  
\E [{\tilde y}_{i^*}] - 3\epsilon_1 - \epsilon_2 \ge 
\E  [z_{i^*}]  - 5\epsilon_1 - 2\epsilon_2,
$$
for $i^* = \arg\max_{i \in [K]} z_i$, 
where the first and last inequalities follow by Eq. \eqref{eq:triangle2}, and the second inequality follows by Eq. \eqref{eq:cs_final_bound}. We have that for any integer $T \ge 2$, and $K = O(\frac{{T^{1/4}}}{\log(T)})$,
$$
\epsilon := 5\epsilon_1 + 2\epsilon_2 \le \frac{20}{\sqrt{\log(T)}} + \sqrt{\frac{2K^2\ln(\sqrt{\log(T)})}{\sqrt{T}}} \le \frac{100}{\sqrt{\log(T)}}.
$$
Thus, by also plugging in the definitions of $z_{\hat{i}}, z_{i^*}$, this concludes the proof of the error bound. The bound on the memory is derived from the guarantee in Theorem \ref{thm:countsketch}, and given by $O(\phi \cdot \frac{1}{\epsilon'^2 \log \frac{T}{\delta'}}) = \tilde{O}(\phi)$, where $\tilde{O}(\cdot)$ hides poly-logarithmic factors in $T$ and $1/\delta$.
\end{proofof}

\begin{proofof}{Theorem \ref{thm:bandit_bai_err_lb}} We will prove a stronger claim, by considering any algorithm that may {\it fully observe} the values of {\it both} arms $i \in [K]$, and is only required to pick a stopping time $t_0$ after which, it returns an arm $\hat{i}\in [K]$ and window size $w$ such that,
\begin{equation}  \label{eq:lb_alg_def}
  \E \left[ \max_{i^*\in[K]}  \frac{1}{w} \sum_{t=t_0}^{t_0+ w} X_{t,i^*} \ - \ \frac{1}{w} \sum_{t=t_0}^{t_0+ w} X_{t,{\hat i}}  \right] \le \epsilon,
\end{equation}
where the expectation is over the randomness of the algorithm.  We will show that any such algorithm must incur expected error $\epsilon \ge 1/8{\log(T)}$.

We now describe a construction of a random instance $X^{2 \times T} \in [0,1]$. Consider a perfect binary tree $\mathcal{T}$ of height $M$, with $2^M$ leaf nodes. We consider functions $f$ which assign a value $f(v) \in [0,1]$ to each node $v \in \mathcal{T}$.  We define a distribution $\mathcal D$ over assignment functions $f$ in a  top–down way as follows. For the root node $v$, its value is always $f(v) = 1/2$, and its sign is $S_v = +1$. 
    For each non-root node $v$ at depth $d\in\{1,\dots,M\}$ we assign it a value $f(v) \in [0,1]$ and sign $S_v \in\{\pm1\}$ at random, based on its parent node sign. Namely, if $p$ is the parent node of $v$, then,
    $$
    S_v =\begin{cases}
			+S_p, & \text{w.p. $\alpha_d$,}\\
            -S_p, & \text{w.p. $1-\alpha_d$,}
		 \end{cases}
    $$
    where $\alpha_d := \frac{1}{2}\left(1 + \sqrt{1-\frac{1}{d}}\right)$. 
Its value is then $f(v) = \frac{1}{2}\left(1 + S_v\sqrt{d/M}\right)$. The randomized instance $X^{2 \times T} \in [0,1]$ is given by the $T=2^M$ leaf node values of two assignments denoted $f_1,f_2$ sampled independently from $\mathcal D$.

We now give the proof of the two simple claims presented in the main paper body, Claim \ref{clm:simple_clm1} and Claim \ref{clm:lb_main_clm}.
\begin{proofof}{Claim \ref{clm:simple_clm1}}
    The proof follows by induction on the depth $d \ge 1$. The base case of $d=1$ follows by definition. 
    For the inductive step, assume the claim holds for all nodes at depth $d-1$, fix a node $v$ at depth $d$. Let $p$ denote its parent, and so $\Pr_{\mathcal{D}}[S_p = +1] = 1/2$. Then, 
\begin{align*}
        \Pr[S_v = +1] &=  \Pr[S_v = +1 | S_p = +1] \cdot \Pr[S_p = +1]  +  \Pr[S_v = +1 | S_p = -1] \cdot \Pr[S_p = -1]  \\
        &=\alpha_d \cdot \frac{1}{2}+  (1-\alpha_d) \cdot \frac{1}{2}  = \frac{1}{2}.
\end{align*}

\end{proofof}

\begin{proofof}{Claim \ref{clm:lb_main_clm}} We can lower bound the probability above as follows,
\begin{align*}
    \Pr & \left[ S_{v_R}^{(1)} \neq S_{v_R}^{(2)} \ \land \ h(S_{v}^{(1)}, S_{v}^{(2)}) \notin \arg\max_{i} S_{v_R}^{(i)}  \right] \\
    &\ge \Pr \left[ S_{v_R}^{(1)} \neq S_{v_R}^{(2)} \ \land \ h(S_{v}^{(1)}, S_{v}^{(2)}) \notin \arg\max_{i} S_{v_R}^{(i)} \ \big| \   S_{v}^{(1)} = S_{v}^{(2)}\right]  \cdot \Pr[S_{v}^{(1)} = S_{v}^{(2)}] \\
    &= \Pr \left[ S_{v_R}^{(1)} \neq S_{v_R}^{(2)} \ \land \ h(S_{v}^{(1)}, S_{v}^{(2)}) \notin \arg\max_{i} S_{v_R}^{(i)} \ \big| \   S_{v}^{(1)} = S_{v}^{(2)}\right]  \cdot  \frac{1}{2}\\
    &= \Pr \left[   h(S_{v}^{(1)}, S_{v}^{(2)}) \notin \arg\max_{i} S_{v_R}^{(i)} \ \big| \ S_{v_R}^{(1)} \neq S_{v_R}^{(2)},  \ S_{v}^{(1)} = S_{v}^{(2)}\right] \cdot \Pr\left[S_{v_R}^{(1)} \neq S_{v_R}^{(2)} \ \big| \   S_{v}^{(1)} = S_{v}^{(2)} \right] \cdot  \frac{1}{2}\\
    &\ge \frac{1}{2} \cdot 2\alpha_d(1-\alpha_d) \cdot \frac{1}{2} = \frac{1}{8d},
\end{align*}
where the first equality follows by Claim \ref{clm:simple_clm1}, and the last inequality by
the fact that conditioning on the event that the parent signs agree, the right-child signs are distributed as two i.i.d Bernoulli random variables, with probability $2\alpha_d(1-\alpha_d)$ of having a different signs. Moreover, for two i.i.d Bernoulli variables, even when conditioning on them disagreeing, the Bayes optimal predictor of which has the higher value  will err with probability $1/2$. 
\end{proofof}

We now continue with the proof of Theorem \ref{thm:bandit_bai_err_lb}, where we will use Claim \ref{clm:lb_main_clm}.
Let $A$ be any algorithm for the simplified task described above \eqref{eq:lb_alg_def}. We first consider any {\it deterministic} algorithm $A$ for the above task, where the observed values are randomized. That is, the algorithm $A$  chooses $t_0, w$, and arm ${\hat i}$ deterministically as a function of the (randomized) values it observes.
Notice that for any choice of $t_0$ and $w$, this determines a parent node $v$ and its two children $v_L$ and $v_R$, such that the algorithm observes all the values $x_1^{(1)},...,x_{t_0}^{(1)}$ and $x_1^{(2)},...,x_{t_0}^{(2)}$, and in particular can deduce both $f_1(v_L)$ and $f_2(v_L)$, and then needs to return $i \in \{1,2\}$ such that $i \in \arg\max_{i^*} f_{i^*}(v_R)$. 
Furthermore, observe that since for any $i \in [K]$, we have that by definition,
\begin{equation}\label{eq:lb_gap}
    |f_{1}(v_R) - f_{2}(v_R)| = \begin{cases}
     \sqrt{d/M}   & \text{ if }S_{v_R}^{(1)} \neq S_{v_R}^{(2)},\vspace{1em} \\  
    0 & \text{ otherwise. }
\end{cases}
\end{equation}

Thus, whenever the two signs of the right-hand-side node $v_R$ differ and the algorithm makes an incorrect guess, it will incur an error of $\ge   \sqrt{d/M}$. Therefore, it remains to lower bound the expected probability of error, 

\begin{align} \label{eq:lb_exp_err}
    \Pr_{f_1,f_2 \sim \mathcal{D}} \left[S_{v_R}^{(1)} \neq S_{v_R}^{(2)} \  \land \  {\hat i} \notin \arg\max_{i} S_{v_R}^{(i)}   \right],
\end{align}
where $v_R, v_L$ are determined by $t_0$ and $w$ chosen by $A$, and ${\hat i}$ is the arm $A$ returned. 
By law of total expectation, it suffices to lower bound the following probability,

\begin{align}
  \Pr_{f_1,f_2 \sim \mathcal{D}} \left[S_{v_R}^{(1)} \neq S_{v_R}^{(2)} \  \land \  {\hat i} \notin \arg\max_{i} S_{v_R}^{(i)} \ \big| \ t_0, w, x_1^{(1)},...,x_{t_0}^{(1)}, \ x_1^{(2)},...,x_{t_0}^{(2)}\right].
\end{align}
Again by law of total probability, we may lower bound the above while also conditioning on the values of the parent signs  $S_{v}^{(1)}, S_{v}^{(2)}$. Moreover, by construction, for any fixed nodes $v, v_R, v_L$, the values of $S_{v_R}^{(1)}, S_{v_R}^{(2)}$ are independent of each other, and only determined by the parent signs, i.e., 

\begin{align}
  \Pr_{f_1,f_2 \sim \mathcal{D}}& \left[S_{v_R}^{(1)} \neq   S_{v_R}^{(2)} \  \land \  {\hat i} \notin \arg\max_{i} S_{v_R}^{(i)} \ \big| \   x_1^{(1)},...,x_{t_0}^{(1)}, 
   \ x_1^{(2)},...,x_{t_0}^{(2)}, S_{v}^{(1)}, S_{v}^{(2)} 
  \right] \\
  & \qquad \qquad = 
  \Pr_{f_1,f_2 \sim \mathcal{D}} \left[S_{v_R}^{(1)} \neq S_{v_R}^{(2)} \  \land \  {\hat i} \notin \arg\max_{i} S_{v_R}^{(i)} \ \big| \  S_{v}^{(1)}, S_{v}^{(2)} \right] \ge 1/(8d), 
\end{align}
where the inequality follows from Claim \ref{clm:lb_main_clm}. Then, by taking expectation over all choices of parent node $v$ (implied by choices of $t_0$ and $w$) that the deterministic algorithm $A$ could have picked based on initial random observations, we get that for any such algorithm, 

\begin{align}
  \Pr_{f_1,f_2 \sim \mathcal{D}} \left[S_{v_R}^{(1)} \neq S_{v_R}^{(2)} \  \land \  {\hat i} \notin \arg\max_{i} S_{v_R}^{(i)} \ \big| \  A \right] \ge \E\left[ \frac{1}{8d} \ \big| \ A \right],
\end{align}
where $d$ is a random variable determined by applying $A$ over the randomized prefix of the data. 
Then, since any randomized algorithm strategy $A$ can be seen as a distribution over deterministic strategies, and since the randomization from $\mathcal{D}$  is independent of the algorithm's randomness, and plugging in the error gap from Eq. \eqref{eq:lb_gap}, we have,
\begin{align}
     \E_A \ \E_{f_1,f_2 \sim \mathcal{D}} \left[ \text{err}(A, X) \right] &=  \E_A \  \E_{f_1,f_2 \sim \mathcal{D}} \left[  \max_{i^*\in[K]}  \frac{1}{w} \sum_{t=t_0}^{t_0+ w} X_{t,i^*} \ - \ \frac{1}{w} \sum_{t=t_0}^{t_0+ w} X_{t,{\hat i}} \ \Big| \ A \right] \\
     &\ge 
\E_A \ \E_{f_1,f_2 \sim \mathcal{D}}\left[ \frac{1}{8d} \cdot \sqrt{\frac{d}{M}}\ \big| \ A \right] \ge \frac{1}{8M}.
\end{align}
In particular, this implies that for any possibly randomized algorithm $A$, there {\it exists} an instance $X \in [0,1]^{2 \times T}$ (determined by  leaf nodes of $f_1,f_2$), for which its expected error is at least ${1/(8M)}$.

\end{proofof}

\begin{proofof}{Theorem \ref{thm:bandit_bai_memory_lb}}
We will prove a stronger claim, by considering any algorithm that may {\it fully observe} the values of {\it both} arms $i \in [K]$, in each round $t=1...T$, and is only required to output the best-arm in hindsight, within its randomly chosen $w$-length window.
We reduce from the well-studied two-party Set-Disjointness (SD) \citep{kalyanasundaram1992probabilistic,bar2004information} on $n=K-1$ elements.
Alice has $A\subseteq[n]$, Bob has $B\subseteq[n]$, and it is promised that either $|A\cap B|=1$ or $|A\cap B|=0$, where the goal is to output  This promise-SD problem has randomized communication complexity $\Omega(n)$. We will show that if $\mathcal{A}$ uses $\sigma$ bits of memory during its run, then it 
can be used to solve the SD problem using $O(\sigma)$ bits of communication, thus proving the lower bound holds. We assume that $\mathcal{A}$ succeeds with constant probability, for concreteness, say probability $2/3$. 

There are $K$ arms: a dummy arm indexed $0$ and $n=K-1$ index arms $1,\dots,n$. 
Draw a public pivot $\tau$ uniformly from $[T]$ (independent of the randomness of $\mathcal A$), and \emph{fix the entire reward sequence before the run} as follows.
For each time $t$ and index arm $i\in[n]$ set
\[
X_{t,i} \;=\; \mathbf 1[t<\tau \ \land \ i\in A] \;+\; \mathbf 1[t\ge \tau \ \land \ i\in B].
\]
Thus, if $i\in A\cap B$ then $X_{t,i}\equiv 1$ for all $t$; if $i\in A\triangle B$ then $X_{t,i}$ is $1$ on exactly one side of $\tau$ and $0$ on the other; otherwise $X_{t,i}\equiv 0$. Then, for the dummy arm, place ones on the central band of length $2\lambda w$ around $\tau$, for $\lambda = 2/5$,
\[
X_{t,0}=\mathbf 1\{\tau-\lambda w \le t < \tau+\lambda w\}.
\]

Let $W=[t_0,t_0+w]$ be $\mathcal A$'s window.   Define the \emph{good-hit} event $\mathsf{Hit}_\lambda$ to be the event that $\tau\in W$ and $W$ contains at least a $\lambda$-fraction of its length on \emph{each} side of the pivot, i.e.
\[
t_0+\lceil \lambda w\rceil \;\le\; \tau \;\le\; t_0+w-\lceil \lambda w\rceil .
\]
For any fixed choice of $t_0$, the number of pivot locations that satisfy the above is exactly $\#\{\text{good }\tau\} \;=\; w - 2\lceil \lambda w\rceil$, as we exclude the first $\lceil \lambda w\rceil$ and last $\lceil \lambda w\rceil$ positions of $W$.
Since $\tau$ is uniform on $[T]$ and independent of $t_0$,
\[
\Pr[\mathsf{Hit}_\lambda \mid t_0] \;=\; \frac{w - 2\lceil \lambda w\rceil}{T-1}
\;\ge\; \frac{(1-2\lambda)w}{T} - \frac{1}{T-1}.
\]
Taking expectation over $t_0$ does not change the bound, so using $w\ge T/c$ we obtain the unconditional constant lower bound
\[
\Pr[\mathsf{Hit}_\lambda] \;\ge\; \frac{1-2\lambda}{c} -  \frac{1}{T-1},
\]
where for fixed $c>1$, $\lambda$, the good-hit event has a positive constant probability. Consider any length-$w$ window $W$ for which the event $\mathsf{Hit}_\lambda$ holds.
By construction, $W$ fully contains the pivot’s central band. Thus, we have the following window-averages for each arm $i \in [K]$:
\begin{itemize}
\item The dummy $i=0$ average is exactly $2\lambda=4/5$ (the central band is fully inside $W$).
\item Any $i\in A\triangle B$ has ones only on one side of $\tau$, hence average $\le 1-\lambda = 3/5$.
\item If $|A\cap B|=1$ with unique $i^\star \in [K]$, then $X_{t,i^\star}\equiv 1$, so its average is $1$.
\end{itemize} 
Therefore, on $\mathsf{Hit}_\lambda$ we have the following strict margins:
\[
\begin{cases}
|A\cap B|=1: & i^\star \text{ beats the dummy by } 1-4/5=1/5,\\
|A\cap B|=0: & \text{dummy beats every index arm by } 4/5-3/5=1/5.
\end{cases}
\]
Since $\varepsilon<1/10$, any $\varepsilon$-correct output on $\mathsf{Hit}_\lambda$ must be $i^\star$ in the intersect case and the dummy in the disjoint case.

Lastly, we simulate $\mathcal A$ as a two-party protocol as follows. Alice (knows $A$) feeds rounds $t<\tau$, sends the $\sigma$-bit memory state to Bob; Bob (knows $B$) feeds $t\ge\tau$ and outputs
``intersect'' iff $\mathcal A$ outputs an index arm, else ``disjoint''.
On $\mathsf{Hit}_\lambda$ the decision is correct, so the protocol has success probability at least $(1-\tfrac13)\cdot(\tfrac{1}{5c}-o(1))$, a positive constant.\\ 

In the above we only guarantee correctness on the “good hit” event $\mathsf{Hit}_\lambda$. To ensure the simulated SD protocol still has success $ > 1/2$, we require that on $\neg\mathsf{Hit}_\lambda$ (which can be verified by an overhead of at most poly-logarithmic communication) Bob outputs a fair coin. Then the success probability is $p \;=\; \Pr[\mathsf{Hit}_\lambda]\cdot\frac{2}{3} \;+\; \bigl(1-\Pr[\mathsf{Hit}_\lambda]\bigr)\cdot\tfrac{1}{2}
\;\ge\; \tfrac{1}{2} + \tfrac{1}{6}\Pr[\mathsf{Hit}_\lambda]
\;>\; \tfrac{1}{2}$, 
since $\mathcal{A}$’s probability of failure is at most $1/3$ and $\Pr[\mathsf{Hit}_\lambda]$ is a fixed constant.  Standard constant-factor amplification—running $O(1)$ independent copies with fresh public pivots and taking a majority—boosts this to success at least $2/3$ while increasing communication by only a constant, thus obtaining the $\Omega(K)$ lower bound.

\end{proofof}

\begin{proofof}{Theorem \ref{thm:bm_mab}}
We start by claiming that there exists a $(\sigma,s)$-memory-bounded learner $\A$ that achieves sublinear regret with a small memory footprint, which in turn will enable us to characterize the behavior of Algorithm \ref{alg:bm_mab}. In particular, for sufficiently large $K$ and $Q$, Algorithm $8$ in \cite{peng2023near} was shown to achieve regret $R_\A(Q) = \tilde O(\sqrt{QK/\sigma})$ with probability at least $1-1/(KT)^{\omega(1)}$
when using $\sigma=O(\text{poly-log}(KQ))$-bits of memory (see Theorem 1.1 in \cite{peng2023near}). At each iteration, the algorithm deals with only a subset (or pool) of experts that is guaranteed to have cardinality $s\leq \sigma$ (as shown in Lemma 4.13 of \cite{peng2023near}). The expert is selected at each iteration by sampling from a distribution $p_\tau$ obtained by normalizing a set of weights that are stored in the algorithm's memory (hence requiring less that $\sigma$ bits) and updated after observing the loss of experts in the pool, as described in the sub-routine of the main algorithm reported in Algorithm 11 of the original paper.

In the following, we study Algorithm \ref{alg:bm_mab} where $\A$ is the $(\sigma,s)$-bounded learner introduced above. We note that without loss of generality we can assume that the loss $\hat c_\tau$ estimated at each iteration of the algorithm to require $\log(T)$ bits. This is because, running $\A$ over the $\log(T)$-truncated values of the observed losses would amount to an approximation error of order $O(\sqrt{T})$ in algorithm's the regret. This implies that constructing and using these losses does not conflict with our memory requirements.

We now apply a reduction strategy from the bandit feedback to the full-information setting, similar to that given  by \cite{blum2007learning} to control the regret of Algorithm \ref{alg:bm_mab}. Specifically, for any block index $\tau$ and corresponding fixed probability $p_\tau$, the block-wise ``exploration'' loss $\hat c_\tau$ is an unbiased estimator of the average loss on $B_\tau$, namely
\begin{equation}\label{eq:c-tau-def}
    \E[\hat c_\tau] = \frac{1}{B_\tau}\sum_{t\in B_\tau} \ell_t =: c_\tau. 
\end{equation}
This follows by construction since the set of exploration times $E_\tau$ is sampled uniformly at random. Moreover, since Algorithm \ref{alg:bm_mab} consists in applying $\A$ to the sequence of losses $\hat c_\tau$, by assumption on $\A$ we have 
\begin{equation}\label{eq:applying-A}    
    \sum_{\tau=1}^Q p_\tau \cdot \hat c_\tau \leq \min_{i\in [K]} \sum_{\tau=1}^Q \hat c_\tau(i) + R_\A(Q),
\end{equation}
with probability at least $1-1/(KT)^{\omega(1)}$, for any sequence of $\hat c_\tau$. 

We can now focus on the loss $\E\left[\sum_{t=1}^T \ell_t(i_t)\right]$ incurred by Algorithm \ref{alg:bm_mab}. Since for every block the algorithm performs always $s'$ exploration steps and since the losses are all upper bounded by $1$, we can control the overall contribution of these terms by $Qs$. We can then ``fill'' in for those missing entries in the loss above, by introducing the expected value of $\ell_t$ when arm $i_t$ is also sampled according to $p_\tau$ (like the other non-exploratory times). This yields the upper bound
\begin{equation}\label{eq:upper-bound-exploratory-times}
\E\left[\sum_{t=1}^T \ell_t(i_t)\right] \leq Qs + \E\left[\sum_{\tau=1}^Q \sum_{t\in B_\tau} p_\tau \cdot \ell_t\right],
\end{equation}
where in the expectation in the right hand side we have moved inside the sum the expectation with respect to sampling $i_t$ according to $p_\tau$ to highlight that there are no exploratory times in there.

Then, by combining \eqref{eq:c-tau-def} and \eqref{eq:applying-A}, we have
\begin{align*}
\E\left[\sum_{\tau=1}^Q \sum_{t\in B_\tau} p_\tau \cdot \ell_t\right] & = |B_\tau|~ \E\left[\sum_{\tau=1}^Q p_\tau \cdot c_\tau \right] \\
& = |B_\tau| ~ \E\left[\sum_{\tau=1}^Q p_\tau \cdot \hat c_\tau\right]\\
& \leq |B_\tau|~\E\left[\min_{i\in[K]} \sum_{\tau=1}^Q \hat c_\tau(i) + R_\A(Q)
\right].
\end{align*}
By the concavity of the minimum and by applying again \eqref{eq:c-tau-def}, we can further observe that
\begin{equation}
    \E\left[\min_{i\in[K]} \sum_{\tau=1}^Q \hat c_\tau(i)\right] \leq 
    \min_{i\in[K]} \E\left[\sum_{\tau=1}^Q \hat c_\tau(i)\right] = \min_{i\in [K]} \sum_{\tau=1}^Q c_\tau(i) = \frac{1}{|B_\tau|} \min_{i\in[K]} \sum_{t=1}^T \ell_t(i), 
\end{equation}
from which we obtain 
\begin{equation}
\E\left[\sum_{t=1}^T \ell_t(i_t)\right] \leq \min_{i\in[K]} \sum_{t=1}^T \ell_t(i) + Qs + \tilde O \left(\frac{T\sqrt{K}}{\sqrt{Q\sigma}} \right).
\end{equation}
Since $Q=T^{2/3}K^{1/3}/\sigma$ by definition and $s\leq\sigma$, the above inequality implies that Algorithm \ref{alg:bm_mab} achieves regret $\tilde O\left(T^{2/3}K^{1/3}\right)$ in expectation.

We now focus on obtaining the same regret bound in high probability. For any $t\in B_\tau$, let $i_t^*$ be a random variable corresponding to $i_t$ for $t\not\in E_\tau$ and sampled according to $p_\tau$ if $t\in E_\tau$ is a stopping time. Then, we have an analogous to \eqref{eq:upper-bound-exploratory-times},
\begin{equation}
    \sum_{t=1}^T \ell_t(i_t) \leq Qs + \sum_{t=1}^T \ell_t(i_t^*).
\end{equation}
Since $\E[\sum_{t=1}^T \ell_t(i_t^*)] = \E\left[\sum_{\tau=1}^Q \sum_{t\in B_\tau} p_\tau \cdot \ell_t\right]$, by applying Azuma-Hoeffding inequality, for any $\delta\in(0,1]$, we have that
\begin{equation}
    \sum_{t=1}^T \ell_t(i_t^*) \leq \E\left[\sum_{\tau=1}^Q \sum_{t\in B_\tau} p_\tau \cdot \ell_t\right] + \sqrt{T\log(1/\delta)},
\end{equation}
holds with probability at least $1-\delta$. By choosing $\delta=1-1/(KT)^{O(1)}$ and applying an  intersection bound to simultaneously control the regret of the $(\sigma,s)$-bounded learner, we have that
\begin{equation}
\sum_{t=1}^T \ell_t(i_t) \leq \min_{i\in[K]} \sum_{t=1}^T \ell_t(i) + \tilde O(T^{2/3}K^{1/3}),
\end{equation}
holds with probability at least $1-\delta -1 /(KT)^{\omega(1)} \geq 1 - 2/(KT)^{O(1)}$, as required.
\end{proofof}

\end{document}